\documentclass[runningheads]{llncs}
\usepackage[T1]{fontenc}
\usepackage{lmodern}
\usepackage{microtype}
\usepackage[subtle]{savetrees}
\usepackage{nicefrac}
\usepackage{makecell}

\usepackage{graphicx}

\usepackage{array}
\usepackage{multirow}

\usepackage{amsmath}
\usepackage{amsfonts}
\usepackage{algorithm}
\usepackage{algpseudocode}

\graphicspath{ {./images/} }

\newcommand{\defaultFigWidth}{1}
\newcommand{\autofigure}[3][\relax]{
	\begin{figure}[h]
		\includegraphics[#2]{#3}
		\caption{#1}
		\label{fig:#3}
	\end{figure}
}

\begin{document}
\title{EDGAR: Embedded Detection of Gunshots by AI in Real-time}
\author{Nathan Morsa\inst{1}\inst{2}\orcidID{0000-0003-4549-7834}}
\authorrunning{N. Morsa}
\institute{Department of Electrical Engineering \& Computer Science, University of Liège \\
\and Research \& Development, FN Herstal, Belgium \\
\email{nathan.morsa@uliege.be}}

\maketitle              %
\begin{abstract}
	Electronic shot counters allow armourers to perform preventive and predictive maintenance based on quantitative measurements, improving reliability, reducing the frequency of accidents, and reducing maintenance costs. To answer a market pressure for both low lead time to market and increased customisation, we aim to solve the shot detection and shot counting problem in a generic way through machine learning. \\
	In this study, we describe a method allowing one to construct a dataset with minimal labelling effort by only requiring the total number of shots fired in a time series. To our knowledge, this is the first study to propose a technique, based on learning from label proportions, that is able to exploit these weak labels to derive an instance-level classifier able to solve the counting problem and the more general discrimination problem.
	We also show that this technique can be deployed in heavily constrained microcontrollers while still providing hard real-time (<100ms) inference.
    We evaluate our technique against a state-of-the-art unsupervised algorithm and show a sizeable improvement, suggesting that the information from the weak labels is successfully leveraged. Finally, we evaluate our technique against human-generated state-of-the-art algorithms and show that it provides comparable performance and significantly outperforms them in some offline and real-world benchmarks.
	
\end{abstract}
\section{Introduction}
\subsection{Motivation for Electronic Shot Counters}
In recent years, the defence industry has seen an increasing interest in preventive and predictive maintenance. In the context of infantry firearms, the number of rounds fired is the prime contributor to their deterioration. Thus, keeping track of the number of rounds fired is an important part of weapon maintenance as it allows a quantitative measure of wear and tear. While this operation has historically been done through logbooks updated manually by operators, these are prone to human error and can prove unreliable. Entries might be omitted or subject to inaccurate estimations. %
The introduction of electronic shot counters to individual weapons allows for much more accurate tracking of weapon usage. These devices are either clipped on or embedded in the firearm and usually rely on MEMS accelerometers, measuring in particular, the acceleration in the firing axis to provide shot detection and counting, burst rate evaluation, and ammunition type discrimination. This allows armourers to perform more accurate maintenance of their inventory by both prioritizing weapons in most need of maintenance and potentially skipping those which have not been operated since their previous maintenance.
In addition, the maintenance can more reliably be performed in a data-driven, predictive manner. Modern firearms maintenance guidelines can be broken down into manufacturer-provided estimations for individual parts or components given as an average lifespan in amounts of shots. This allows the armourer to replace those parts preemptively, avoiding potential firearms malfunction during operation.
This change in maintenance paradigm is estimated to save up to 50\% on armourer labour time, increase weapon availability up to 90\%, and reduce operating costs by up to 20\%.~\cite{fnherstalSAM2020}

While some shot counters employ different types of sensors, our proposed method should be equally applicable to any temporal data from one or a combination of sensors. Although microphones are a popular method of shot counting in controlled environments such as shooting ranges, they require exposing an external sensor that is not compatible with military requirements for weapons which include prolonged exposure to hostile environments such as water, salt, dust, grease or acid. They also suffer from echoes in enclosed environments requiring frequent recalibration. Magnet-based methods have shown very effective; however, the requirement for close proximity to the firing mechanism is often a cause for concern due to either encumbrance or safety reasons. They also require more expensive parts. As a result, accelerometer-based solutions have been preferred by the defence market as these can be fully and invisibly embedded in any part of the weapon, in particular grips and handles, which are often hollow and separated from the firing mechanism.

\subsection{Counting Problem}
The problem of counting the number of shots in a time series lies firstly in detecting relatively-rare individual events from unrelated ones, such as normal weapon manipulations or falls on hard ground interspersed in-between shots. We illustrate some example inputs in Figure~\ref{fig:graph_paper_demo_mosaic}. Example \#1 shows a shot that is followed in close proximity by a purely mechanical event which should not be counted. Example \#2 illustrate other non-shot events. In addition, shots need to be discerned from each other for proper counting and the start of each one properly identified for burst rate evaluation. However, shots can present a wide variety of signatures depending on external factors (see Section~\ref{external_variables}). Example \#3 shows shots fired with the same weapon as \#1 presenting differing signatures both from \#1 and from each other even though they happen in close succession. Example \#4 shows how shots can blend into each other rendering individual detection difficult, with the common occurrence of a mechanical event also blending in at the end of a burst. Mechanical events visually very similar to shots also happen in close proximity. Examples \#4, 5 and 6 are taken from different weapons and show how the signature can significantly differ between them.

\autofigure[Example inputs (high resolution available in digital version). Events have been individually labelled by a human expert between shot in green (numbered at the bottom) and non-shot in red.]{width=\defaultFigWidth\textwidth}{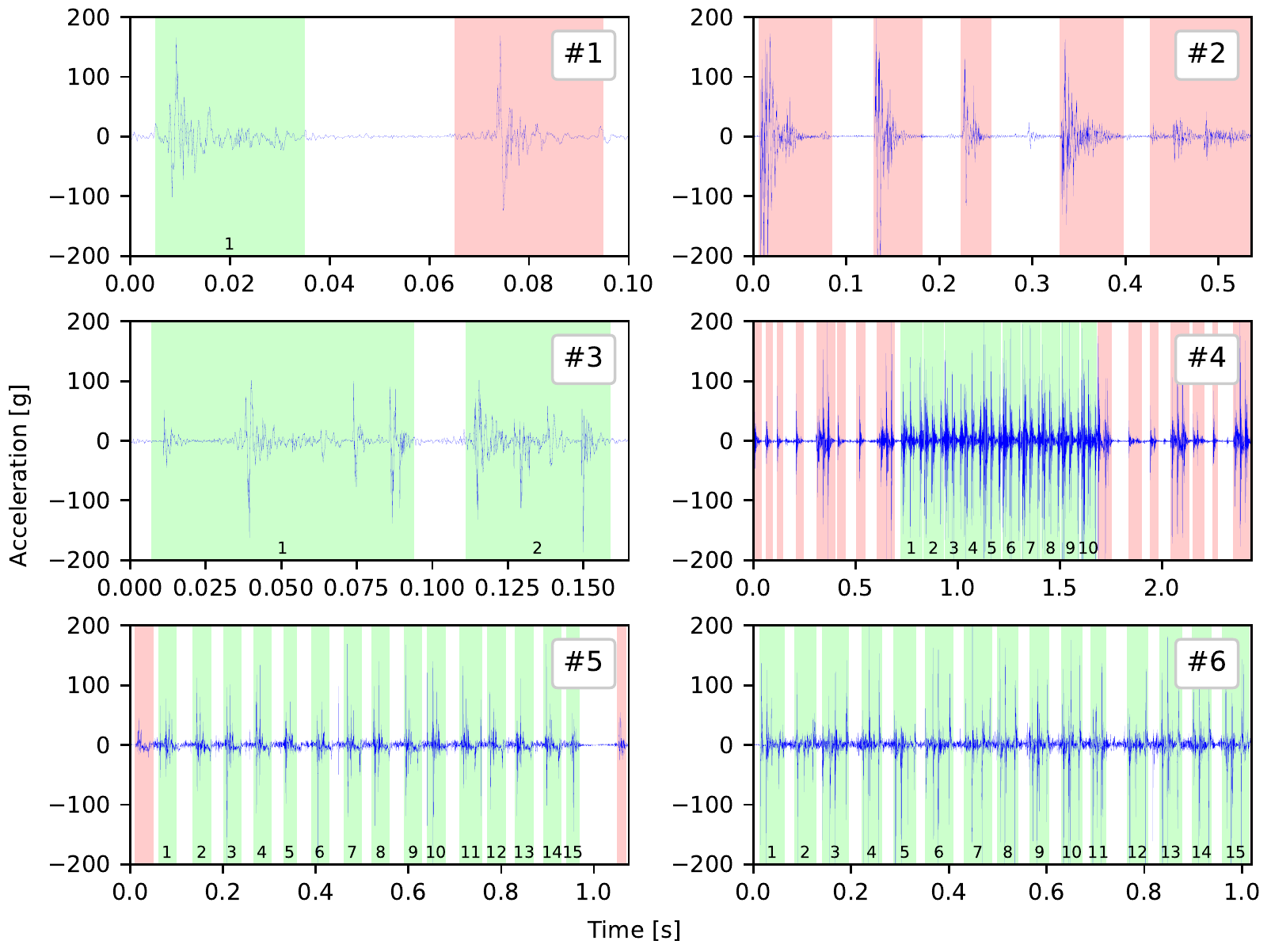}

\subsection{The Case for Machine Learning}

As the market for shot counters develops, requests for new types of shot counters have increasingly high requirements and lower accepted lead time. %
Nowadays, clients of the defence industry expect a higher grade of personalisation, including modifications impacting the core design of the weapon and necessitating new R\&D work. On the other hand, fast product delivery is also more and more expected in an increasingly competitive market. This has led to a fundamental restructuring of the manufacturing processes from make-to-stock to assemble-to-order and even engineer-to-order strategies.

In recent public tenders of the firearms market, shot counters have progressively become an important criterion in the attribution process. It is thus growingly important for shot counters to be available with very low delay, ideally at the latest when the weapon prototype is delivered to the client for initial testing so that the shot counter can be evaluated alongside its platform. However, since a shot counter can only be developed when at least one corresponding weapon design prototype is available, the window of time left for development between the end of the weapon R\&D work and the first delivery can be slim, sometimes in the order of a few weeks.
Combined with the increasing complexity of the many existing weapon variants leading to a correspondingly high required development time as described in Section~\ref{long_dev_time}, the current techniques of manually written algorithms can result in a failure to meet the market.

The automatisation of the shot counting algorithm generation aims to alleviate this issue by trading labour time from a qualified expert to computation time, which can be scaled through cloud computing according to urgency. In addition, automatically-generated algorithms have been shown to outperform and replace human-generated ones in some situations.
However, as further described in Section~\ref{related_work}, there is currently no publicly available generic technique for automatically deriving shot counting algorithms.

\subsection{Contributions}
Our paper makes the following contributions:
\begin{itemize}
	\item We describe a procedure allowing one to use a single-axis accelerometer to construct a dataset representative of a firearm's behaviour with minimal work related to data labelling through weak labels only.
	\item We propose a new technique that is able to exploit these weak labels to derive an instance-level classifier able to solve the counting problem and the more general discrimination problem.
	\item To our knowledge, we are the first to propose a neural network structure suitable to solve this problem in real-time on %
	embedded microcontrollers.
	\item We propose a series of generic and domain-specific improvements to the base technique allowing it to reach much higher performance levels.
	\item We evaluate our technique against a state-of-the-art unsupervised algorithm and show a very large improvement, suggesting that the information from the weak labels is successfully leveraged.
	\item We finally evaluate our technique against human-generated state-of-the-art algorithms and show that it provides comparable performance and significantly outperforms them in some offline and real-world benchmarks.
\end{itemize}

\section{Related Work} \label{related_work}
Owing to the specific nature of the problem and the relatively recent industry interest in embedded shot counters, publicly available academic research on this specific problem is scarce. FN Herstal owns several international patents on the topic since 2006. In particular, one covers the use of successive events in accelerometer data for the purpose of shot counting~\cite{joannesDeviceDetectingCounting2010} and has likely hindered further research into exploiting this signal. %
Loeffler~\cite{loefflerDetectingGunshotsUsing2014} and Reese~\cite{reeseSituationalawarenessSystemNetworked2016} limit themselves to low scale and sampling frequency, are restricted to a few shots and do not make use of machine learning. Ufer et al.~\cite{uferSelfCalibratingWeapon2014} offer a calibration technique on a pre-made expert algorithm. Calhoun et al.~\cite{shotspotterpatent} propose a method for gunshot detection involving deep learning; however, this technique is aimed solely at acoustic detection (not counting) from a network of city-wide microphones and involves a human in the loop.

Inspiration could be gleaned from the richly studied domain of fall detection. For example, Putra et al.~\cite{putraEventTriggeredMachineLearning2017} study accelerometer-based fall detection by decomposing the impact into sub-events in the time domain, thereby showing similarities with our problem and applying machine learning techniques. A recent paper by Santos et al.~\cite{santosAccelerometerBasedHumanFall2019} applies deep-learning techniques to this same problem and proposes a CNN with good results when using data augmentation techniques. These approaches however rely on strongly labelled data.

A recent patent application by Weiss et al.~\cite{weissDeviceSystemMethod2021} for Secubit Ltd. details concurrent work aimed at using deep-learning %
for shot counting from %
accelerometer data, facultatively augmented by other sensors.%
The method proposed in this work differs from the Weiss et al. patent in many aspects:
\begin{itemize}
	\item A concrete proposition is made for the model structure, data preprocessing, effective input vector selection, dataset construction, training process and an effective method for ammunition type discrimination. %
	\item Our work proposes an innovative technique allowing us to leverage the low-effort data about the number of shots contained in a time series. This allows efficient learning directly from the input dataset in its whole variance without the need for potentially unrepresentative GAN-synthesized data.%
	\item Our technique shows performance best with an input vector significantly smaller than the normal event length.%
	\item Our %
	technique can be entirely performed in an embedded device in real-time, requiring no intermediate representation to be stored and/or transferred. %
	\item The Weiss et al. proposition does not make a concrete description of how their technique can be employed in an embedded device, apart from the flow control system and mentioning possible optimizations through graph pruning and knowledge distillation. Our technique bypasses the need for these optimizations by directly training a network adapted to the target platform. %
\end{itemize} 

\section{Dataset Acquisition and Construction}

For a given model and calibre of firearm, the following external variables are known to have a significant impact on the weapon behaviour: shooting sequence, ammunition type, ammunition load, gas-operated reloading nozzle size, shooter position/mounting mechanism, mounted accessories weight, ammunition loading type, usage of a suppressor, shooting angle, weapon and canon temperature, firearm dirtiness, and firearm wear and tear (by including both new and used weapons).

\label{external_variables}

\label{shooting_plan_reason}
To acquire a dataset representative of the whole spectrum of possible real-life weapon behaviours, one would ideally have to record and control these external variables independently. However, the number of %
variables and the large space of possible values for each of them can rapidly lead to an unmanageable number of combinations. As a result, a discrete and possibly reduced number of significantly differing settings for each variable will be chosen according to expert armourer knowledge of the weapon and existing experimental results. A reduced number of combinations will then be chosen according to the development budget, attempting to capture both nominal and extreme behaviours of the weapon.

\label{weak_labels}
Raw data can be acquired by recording the sensor output while operating the weapon. However, the labelling of this data presents a significant challenge: Creating strong labels identifying the position of individual shots in the data requires the intervention of a firearms expert to discern shots from unrelated acceleration events. This is a very labour-intensive task and is prone to human error. On the other hand, weak labels in the form of a total number of shots fired in a given recording can be produced cheaply and reliably by manually counting the number of rounds used, especially when these come in pre-numbered containers such as ammunition boxes and magazines.

\label{long_dev_time}

\section{Proposed Method} \label{EDGAR}

To exploit our dataset, we need to find a technique allowing us to work from the weak labels. This can be accomplished by finding a way to reformulate the counting problem into a category proportion problem. To do so while keeping a detector with suitable time and space constraints for embedded use, two hypotheses are made:

\begin{enumerate}
	\item A shot has a finite and known maximum duration.
	\item At least one sub-event of a shooting event can always be reliably distinguished from the background noise.
\end{enumerate}

Hypothesis 1 is related to the length of the candidate windows that will be considered. Since it imposes no maximum bound on the duration, a sufficiently large number will always exist to satisfy it. However, larger numbers will negatively impact the performance of the resulting detector. A good number can be easily derived from the minimum theoretical burst rate of the weapon.

Hypothesis 2 allows us to reduce the number of candidate windows that will be considered. It is done by producing candidate windows only when a certain metric is satisfied, preventing useless computations during rest periods where the only input is background noise. The difference between metric and detector is that the metric is not subject to any constraint on the number of false positives it provides. However, a metric with fewer false positives will further reduce the computation time. It is important that the metric avoids false negatives, which would result in a valid candidate not being presented to the detector. %

\label{metric_llp}
Satisfying hypothesis 2 in practice will depend on the nature of the input signal. For the most common input signal of accelerometer time series, we propose the use of a rolling average on the instantaneous accelerations squared:

\begin{equation} \label{eqn:preprocess_metric}
	m[t] = \dfrac{1}{w} \sum_{i = -w/2}^{w/2} (a[t+i+o])^2
\end{equation}

Where $a[t]$ is the input acceleration time series, $w$ is a positive integer hyperparameter for the size of the metric window, and $o$ is an optional integer offset that can be applied to shift the position of the metric relative to the input signal.%

The input signal and metric are illustrated in Figure~\ref{fig:example_input_compressed}. It can then be compared against the high ($T_H$) and low ($T_L$) thresholds, whose values are hyperparameters of the model. A candidate would, for example, only be generated when the signal dipped below $T_L$ and increased above $T_H$.

\autofigure[Example of candidate generation on an ideal input.]{width=\defaultFigWidth\textwidth}{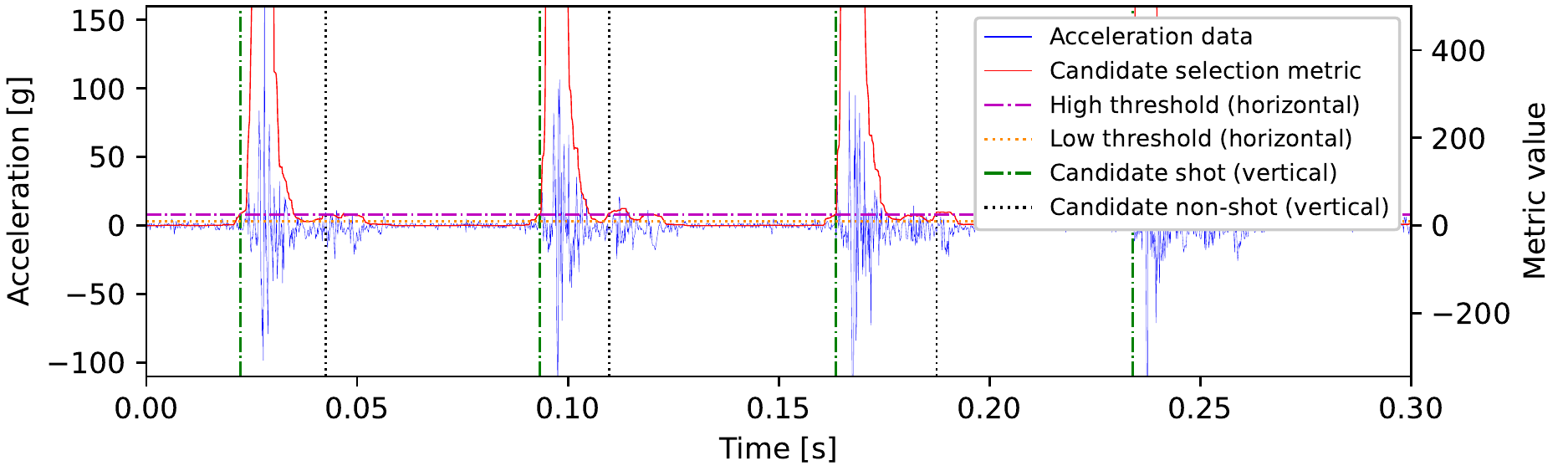}

Each candidate defines a slice of the input series of a fixed size determined according to Hypothesis 1 at most. In practice, experiments have shown that similar or better classification performance can be obtained with a significantly smaller input vector. A smaller input vector has a major benefit regarding inference-time performance.

\label{multi_candidates}

Let $N$ be the number of categories of inputs to classify. The classification categories can be arbitrarily decided as long as their number of occurrences can be known for each input series. In the simplest case, a database could be classified as \textit{non-shot} or \textit{shot}, leading to $N=2$. A more complex example would be classifying between \textit{non-shot}, \textit{shot with live ammunition}, and \textit{shot with training ammunition} leading to $N=3$. Shot types could then be further subdivided according to whether or not they include a suppressor with $N=5$.
Given a time series including a shooting sequence with different counts $\mathbf{c}_i$ with $i \in \{1,...,N\}$ being the number of occurrences of each event category. Let its division into candidate slices be $\mathbb{X}$. We can then define event proportions between the known number of events of a certain category and the total number of candidates as %
	$\mathbf{p}_i = \frac{\mathbf{c}_i}{|\mathbb{X}|}$.
A machine-learning-based classifier $f_\theta(\mathbf{x})$ with trainable parameters $\theta$ can then be run on each candidate input vector, attempting to classify it into its corresponding category. The resulting predictions over a given time series can then be aggregated into proportions as follows:
\begin{equation} \label{eqn:proportion_def}
	\hat{\mathbf{p}} = \frac{1}{|\mathbb{X}|} \sum_{\mathbf{x} \in \mathbb{X}} f_\theta(\mathbf{x})
\end{equation}

The two proportions can then be compared according to a label proportion technique such as the one described by Tsai et al.~\cite{tsaiLearningLabelProportions2019} to tune the trainable parameters $\theta$ to optimal values. In particular, the related part of the loss function will be:

\begin{equation} \label{eqn:lprop}
	\mathcal{L}_{prop} = -\sum_{i=1}^{N}\mathbf{p}_i log(\hat{\mathbf{p}}_i) +\sum_{i=1}^{N}\mathbf{p}_i log(\mathbf{p}_i)
\end{equation}

The second term is an original addition allowing the loss function to converge to 0 in the case of perfect predictions (i.e., when $\hat{\mathbf{p}} = \mathbf{p}$). Without it, the minimal achievable loss has different values depending on the ratio of the different proportions. This makes the comparison of loss values across different inputs more meaningful, as well as their aggregation in an average loss over the dataset. Average loss can be viewed as a smooth metric of model quality, whereas the model accuracy can only take discrete values, corresponding to a discrete number of counting errors. This enables us to monitor the model learning evolution more precisely and employ early stopping techniques based on the average loss on the validation dataset. Comparative experiments have shown that its addition provides increased numerical stability and prediction accuracy.

In theory, this formulation does not guarantee the expected assignation of predicted classes in the edge case where the proportion of some of them are systematically equal ($\mathbf{p}_i = \mathbf{p}_j$) in every sample of the training set. However, this is not a practical concern as the natural variability in the number of occurrences of each class will quickly break any potential tie and allow the model to converge appropriately. In addition, it is generally easy to obtain samples containing only non-shot data or a single class of shot data thereby providing some heavily skewed samples preventing this problem from occurring.

This technique allows us to train a model from our weak aggregated labels. Note that our proposed method only relies on a particular definition of the loss function while not imposing any constraints on the type or structure of the model. A proposed model for embedded usage is described in Section~\ref{model_structure}, but can be swapped for other types of neural networks or machine learning techniques as long as it is able to train from such a loss function. \label{any_model}

\label{physical_constraints}
\subsection{Minimum Cycle Time} \label{minimum_cycle_time}
An important factor limiting the performance of the technique, as previously described, appears when the metric generates two candidates very close in time. The model will receive two very similar input vectors, different slices of the same shot, which it is likely to both classify as being a shot, leading to false positives in the counting as a typical firearm is only able to shoot one projectile at a time in a cycle. However, we could leverage the known information of the minimum cycle time of the weapon to alleviate these false positives. For example, the modern Minimi 5.56 specifications allow a maximum firing rate of 950 rounds per minute (rpm), corresponding to a 63 ms average cycle time. Since this burst rate can be slightly exceeded in exceptional situations and individual cycle times might vary, it is necessary to incorporate a margin of error over that theoretical number. A value of 40 ms (corresponding to a theoretical cycle rate of 1500 rpm) has been used in our experiments.

While we could consider filtering out candidates at the metric level, at this step, we do not yet know which candidate the model would usually classify as a shot. Doing so thus risks removing the candidate closest to the actual shot ignition in favour of one before or after it. This could lead to non-detections and would diminish the model's ability to precisely identify the position of the shot in the time series. A better approach is to apply this filtering after the network predictions. We first change successive \textit{shot} predictions to \textit{non-shot} if they fall within an exclusion window of the first prediction. Experimental data (see Section~\ref{performance_breakdown}) show that this already leads to a significant reduction in error rate. In addition, we can avoid performing model inference on the ignored predictions leading to an overall computational performance increase in deployment.
A drawback of implementing this as a post-processing step is that the model has no knowledge of it and remains penalized during training for the duplicate predictions. This might produce a model overall unnecessarily "reluctant" to predict shots. In other words, one can assume that if the model had knowledge of this post-filtering, it would not hesitate to predict a shot for slightly offset candidates, knowing that it would not be penalized for duplicate predictions. We can accomplish this during training by implementing Algorithm~\ref{alg:prediction_mask}, which corrects the predictions while ensuring the duplicates do not take part in the loss computation. This process is performed in an iterative fashion since a masked-out shot prediction must not itself create a mask. Redefining Equation~\ref{eqn:proportion_def} to substitute $f_\theta(\mathbf{x})$ by $f'_\theta(\mathbf{x})$ will then prevent backpropagation for the masked-out predictions. This leads to another significant reduction in error rate in experiments.

\begin{algorithm}[H]
	\caption{Remove duplicate predictions in training}\label{alg:prediction_mask}
	\begin{algorithmic}
		\Require $t_i$: timestamp for each candidate $\mathbf{x}_i$.
		\Require $T_M$: minimum event duration	
		\Require $f_\theta(\mathbf{x})$: instance-level classifier with trainable parameters $\theta$,
		\Require $\mathbf{e}$: vector corresponding to a non-shot prediction with maximum certainty
		\State $\hat{\mathbf{y}}_i \gets \text{max}_{1 \leq k \leq N} f_\theta(\mathbf{x})_{i,k}$  \Comment{Individual category predictions}
		\State $\mathbf{m}_i \gets \top$ \Comment{Mask to be computed}
		\For{$i = 1,...,|\mathbf{x}|$}
		\State $\hat{\mathbf{y}}'_j \gets
		\begin{cases}
			\hat{\mathbf{y}}_j,  & \text{if } \mathbf{m}_j\\
			0,              & \text{if } \lnot \mathbf{m}_j
		\end{cases}, j \in \{1,...,|\mathbf{x}|\}$
		\If{$\hat{\mathbf{y}}'_i$ predicts a shot}
		\State $\mathbf{update}_j \gets (t_j \leq t_i) \lor (t_j > t_i + T_M), j \in \{1,...,|\mathbf{x}|\}$
		\State $\mathbf{m} \gets \mathbf{m} \land \mathbf{update}$
		\EndIf
		\EndFor
		\State  $f'_\theta(\mathbf{x})_j \gets
		\begin{cases}
			f_\theta(\mathbf{x})_j,  & \text{if } \mathbf{m}_j\\
			\mathbf{e},        & \text{if } \lnot \mathbf{m}_j
		\end{cases}, j \in \{1,...,|\mathbf{x}|\}$
		\State \Return $f'_\theta(\mathbf{x})$
	\end{algorithmic}
\end{algorithm}

\subsection{Model Structure} \label{model_structure}
We have chosen to implement our model as a convolutional neural network (CNN) inspired by the one described by Santos et al.~\cite{santosAccelerometerBasedHumanFall2019}, as described in Figure~\ref{fig:network_compressed_vector}. The illustrated model totals only 33242 parameters, as we look for a small model suitable for real-time embedded inference.

\label{network_structure}
\autofigure[Neural network structure for an input size of 232. Convolutions do not use strides nor padding. The input size and number of convolution channels depends on the application and the available computational budget.]{trim={1.05cm 21.5cm 1.40cm 1.55cm}, clip,width=\defaultFigWidth\textwidth}{\detokenize{network_compressed_vector}}

The original ReLU activations have been changed to ReLU6 to make the network more suitable for quantization. ReLU6 activation bounds the values, limiting their possible range. This allows us to use a fixed point representation with more bits allowed to the fractional part, reducing the quantization artefacts and improving overall accuracy. Results outside the scope of this paper have shown the difference in error rate before and after quantization to be reduced both in median and variance when applying ReLU6 activations.

In testing, the maximum value of 6 has shown to produce the best median results, whereas ReLU2 has shown the best best-case performance. Since 6 is not a power of 2, the full $[0;6]$ range does not quantize efficiently, needing up to 3 bits for the integer part while not making full use of them. However, we speculate that even when the distribution of activation values uses the full range, only a low percentage of values actually falls near the bounds. Better performance can thus be achieved in those cases by only quantizing a 2-bit range around the average, saturating the outliers but leaving one more bit to be used in the fractional part.

As previously mentioned, this paper focuses on the definition of a loss function allowing the leveraging of our weak labels independently of the trained model type. This network serves as a basis proving the viability of the technique on heavily constrained hardware, and the derivation of the optimal model structure will be the subject of future research. While our small model and dataset sizes allow for relatively fast training, we plan on exploring how MINIROCKET~\cite{dempsterMINIROCKETVeryFast2021} could enable us to further reduce it thereby allowing a faster exploration of the hyperparameter space. With proper hardware support, spiking neural networks could also be a good candidate for this type of application. While this work focuses only on instance-level information, long short-term memory (LSTM) networks could prove a worthwhile extension to add contextual information to the prediction.

\subsection{VAT} \label{VAT}
Following the good results of Tsai et al.~\cite{tsaiLearningLabelProportions2019}, we apply Virtual Adversarial Training (VAT) as described by Miyato et al.~\cite{miyatoVirtualAdversarialTraining2018}. As advised in the original paper, optimization is only done through the perturbation size $\epsilon$. The regularization coefficient $\alpha$ is fixed at $1$, the finite difference factor $\xi$ is fixed at $10^{-6}$, and a single power iteration is performed ($K=1$).
Contrary to the findings of Laine et al.~\cite{laineTemporalEnsemblingSemiSupervised2017}, our experiments have shown that the best results on our problem are obtained when the VAT loss is introduced as soon as possible. Thus, we do not include any ramping up to the VAT loss component.

\section{Experiments}
\subsection{Methodology}
We chose to evaluate the technique on two firearms: the FN Minimi\textsuperscript{\tiny\textregistered} 5.56 %
and the FN\textsuperscript{\tiny\textregistered} M2HB-QCB%
. The Minimi was selected due both to its availability for in-depth testing and its reputation of being a notoriously difficult weapon to provide counting algorithms for due to its very wide number of possible configurations. In addition, the weapon tends to show low-information signals in some configurations due to its heavy weight regarding the low-power 5.56 ammunition. The M2 was selected as a contrast to the Minimi, being on the higher end of ammunition power and using a completely different action mechanism.

External variables considered for the Minimi were: ammunition type (live/ blank), barrel size (short/long), accessory weight (none/3kg accessories), gas-operated reloading nozzle size (minimal/nominal/maximal), shooter position (shoulder/bipod/waist), and firing sequence (semi-automatic, three to six rounds bursts, 4-1-4-1 bursts, full bursts). Data were acquired in groups of $\sim$15 rounds. Non-shot data acquisition includes: dry firing, 1.5m falls onto concrete, opening and closing the top cover, full reloading manipulation, and user randomly bumping the weapon. In addition, similar data were also acquired on the Minimi~7.62.

External variables considered for the M2 were: ammunition type (live/blank), the weight of ammunition belt (minimal belt/100 rounds belt), mount (tripod, fixed mount, elastic mount, deFNder\textsuperscript{\tiny\textregistered} teleoperated station), and firing sequence (manual rearming, single rounds in automatic mode, 3-6 rounds bursts, 4-1-4-1 bursts, full bursts). Data were acquired in groups of $\sim$10 rounds. Non-shot data acquisition includes: dry firing, rearming and releasing the mobile parts, opening and closing the top cover, and user randomly bumping the weapon.

In all cases, input that contains firing incidents (weapon malfunctions) is rejected. To evaluate the final performance of the algorithm, a validation dataset is split from the input dataset. To ensure that the validation set captures the behaviour of the firearm in a wide range of situations, the input series are first sorted into different bins for each available combination of external variables. The validation set is then constructed by randomly sampling 10\% of the data in each bin, rounding in favour of the validation set. The composition of the different datasets is shown in Table~\ref{tab:datasets_shots}. Note that the number of non-shot events is not exactly known. This is firstly because there is no easy and reliable way to provide weak labels for these data, meaning we have to rely on operator estimations. In addition, a significant number of non-shot events is also acquired as part normal shooting tests as the operator has to manipulate the weapon before and after firing sequences for both practical and safety purposes. These are unaccounted for in our count. Thus, we only provide a lower bound on the number of non-shot events. A measure of their actual number can be gained from the total number of candidates generated by the preprocessing.

The real-world distribution between shots and non-shots is also unknown, as this will heavily depend on the end user and their doctrine. For example, some users will frequently manipulate the weapon without shooting during training and/or perform dry firings and safety checks before or after shooting, and transport it in vehicles generating significant vibrations. Other users will leave the weapon mostly in storage, and do a minimal number of manipulations around the shooting. We thus evaluate these separately, by including in the firing samples only the minimal number of manipulations and safety checks around the shooting (which also generates unavoidable non-shot candidates due to the inherent behaviour of the weapon). Other non-shot events are ideally sampled separately so that a measure of the number of false positives generated during weapon manipulation can be given. These will be given in the "non-shot only" rows of the following section.
\begin{table}
	\caption{Summary of the number of shots per dataset.}\label{tab:datasets_shots}
	\begin{tabular}
		{
			>{\raggedleft}p{0.11\textwidth}
			>{\centering}p{0.13\textwidth} %
			>{\centering}p{0.13\textwidth}
			p{0.01\textwidth}
			>{\centering}p{0.13\textwidth} %
			>{\centering}p{0.13\textwidth}
			p{0.01\textwidth}
			>{\centering}p{0.13\textwidth} %
			>{\centering\arraybackslash}p{0.14\textwidth}
		}
		\hline
		\multirow{2}{*}{}
		&\multicolumn{2}{c}{Minimi 5.56}
		&&\multicolumn{2}{c}{Minimi 7.62}
		&&\multicolumn{2}{c}{M2}
		\\ \cline{2-3} \cline{5-6} \cline{8-9}
		&Learning&Validation&&Learning&Validation&&Learning&Validation\\
		Live & 4461 & 1785 && 4707 & 1800 && 5263 & 729\\
		Blank & 2130 & 719 && 943 & 348 && 4238 & 529\\
		Non-shot events & >400 & >55 && >97 & >35 && >1550 & >180\\
		Candidates (total) & 14857 & 5494 && 12029 & 2355 && 60844 & 8096
	\end{tabular}
\end{table}

Randomly initialized neural networks are then trained in groups of 20. Optimization is done through stochastic gradient descent with fixed $0.9$ Nesterov momentum. The learning rate is reduced by a factor of 2 for every 20 epochs without improvement larger than $10^{-5}$ on the validation loss. A learning phase is stopped if 40 epochs occur without any improvement on the validation loss. Learning occurs in three phases: an optional pre-training phase on a wider dataset, normal training on desired firearm's dataset, and quantization-aware training on the same dataset. After each phase, the best model is selected according to the lowest number of validation errors, then the lowest validation loss in case of a tie and proceeds to the next phase or final evaluation. The resulting network is quantized to eight bits and runs on the target platform through the TFLite for Microcontrollers framework accelerated by the CMSIS-NN~\cite{laiCMSISNNEfficientNeural2018} library.

To iteratively optimize the different hyperparameters, we start by optimizing individually the preprocessing hyperparameters starting with $w$ from Equation~\ref{eqn:preprocess_metric}, the associated thresholds $T_H$ and $T_L$ and the network input vector length. We then proceed to optimize the learning rate and VAT perturbation size $\epsilon$. The model can then be reduced according to the desired performance/computational budget tradeoff by reducing the number of convolution channels. A second round of fine-tuning can then be applied. The model shows a single minimal error rate for all hyperparameters, which can thus be optimized by bisection until the minimum is found. The only exception is $w$ which has been shown to produce several local minima requiring a more thorough exploration within the acceptable computational budget. The details of the hyperparameter survey are outside the scope of this article and will be included in a future paper. The values used for our experiments are reports in Table~\ref{tab:hyperparams_EDGAR}.

\begin{table}[!ht]
	\centering
	\caption{Chosen preprocessing and EDGAR hyperparameters.} \label{tab:hyperparams_EDGAR}
	\begin{tabular}
		{
			>{\centering}m{0.2\textwidth}
			||>{\centering}m{0.15\textwidth}
			|>{\centering}m{0.15\textwidth}
			|>{\centering\arraybackslash}m{0.15\textwidth}
		}
		Hyperparameter & \makecell{Minimi 5.56\\(\#1)} & \makecell{Minimi 5.56\\(future)} & M2 \\ \hline
		$|x|$ & 232 & 232 & 360 \\
		$w$   & 5ms & 5ms & 5ms \\
		$T_H$ & 30  & 114 & 114 \\
		$T_L$ & 10  & 90 & 90 \\  \hline
		Filters & 18 & 18 & 64 \\
		Learning rate & 0.002 & 0.002 & 0.0032 \\
		VAT $\epsilon$ & 5 & 5 & 5
	\end{tabular}
\end{table}

For the Minimi, pre-training is performed on the 5.56 and 7.62 datasets. For the M2, pre-training is performed on all three datasets. Acquisition and testing are done on a custom hardware platform, including a 64MHz Cortex-M4F microcontroller and a $\pm$200g, 6400Hz MEMS accelerometer.

Due to technical limitations of our current data collection setup, the extraction of samples currently requires the intervention of a trained technician on the shooting range making it significantly more expensive than autonomous counting operation. As a result, we have chosen to test the final performance of our method in this fashion pending the acquisition of truly independent test datasets.

As detailed in section~\ref{related_work}, no ML-based approach or device is currently publicly available for our problem. While the domain of time series classification is well studied, as described in section~\ref{weak_labels} the lack of labels for individual shots prevents us from using supervised techniques. As a baseline, we include for comparison a method of unsupervised clustering on the candidates based on deep learning: Deep Temporal Clustering (DTC) as described by Madiraju et al.~\cite{madirajuDeepTemporalClustering2018}. Hyperparameters have been fine-tuned following a similar procedure and are reported in Table~\ref{tab:hyperparams_DTC}, and pre-training is also applied. %
\begin{table}[!ht]
	\centering
	\caption{Chosen preprocessing and DTC hyperparameters.} \label{tab:hyperparams_DTC}
	\begin{tabular}
		{
			>{\centering}m{0.2\textwidth}
			||>{\centering}m{0.15\textwidth}
			|>{\centering\arraybackslash}m{0.15\textwidth}
		}
		Hyperparameter & Minimi 5.56 & M2 \\ \hline
		$|x|$ & 232 & 360 \\
		$w$ & 5ms & 5ms \\
		$T_H$ & 126 & 114 \\
		$T_L$ & 119 & 90 \\  \hline
		Distance metric & CID & CID \\\
		$\alpha$ & 17.5 & 1 \\
		$\gamma$ & 90 & 1 \\
		Batch size & 64 & 64 \\ 
		Pool size & 4 & 8 \\ 
		Kernel size & 96 & 10 \\ 
		Filters & 50 & 50 \\ 
	\end{tabular}
\end{table}

For the human baseline, we compare our technique with FN SmartCore\textsuperscript{\tiny\textregistered} shot counters, which share the same hardware and sensor, running algorithms generated by human firearm experts. These work by first detecting discrete shocks in the input signal%
. They then attempt to interpret groups of shocks as sub-events of a firing cycle (such as feeding, locking, firing or rearming) by considering their relative energies, timings, directions and the duration of "calm zones" in-between~\cite{joannesDeviceDetectingCounting2010}. These devices have been commercially available since 2012 %
and are available for a wide range of machine guns and assault rifles.

\subsection{Performance}
In this section, we examine the performance of our technique on both the validation datasets and real-world testing. Since we are dealing with weak labels, we are unable to verify model predictions at the instance level. Models are evaluated through their error rate, which we define as the sum of the counting differences (category by category) for all time series, divided by the real total count. More formally, we define the error rate $E$ as:

\begin{equation}
	E = \dfrac {\sum_{i,j}{\left| \hat{\mathbf{c}}_{ij} - \mathbf{c}_{ij} \right|}}
	           {\sum_{i,j}{\mathbf{c}_{ij}}}
\end{equation}
where $c_{ij}$ is the count of shots (or, if applicable, other countable events) of type $i$ for time series $j$ and $\hat{\mathbf{c}}_{ij}$ is the one estimated by the model.

A model that always predicts non-shots will have an error rate of 100\% by this definition. In practical applications, non-shots constitute the majority of candidates (see Table~\ref{tab:datasets_shots}). In these conditions, a model that always predicts shots will have an error rate over 100\% (\textbf{119\%} for the Minimi 5.56 and \textbf{544\%} for the M2 datasets). We can expect applications with a larger class imbalance between non-shots and shots to show a higher base error rate due to the increased number of false positives, denoting a harder problem for shot counters. Note that we could also define the error rate compared to the total number of candidates, resulting in smaller error rates but with identical relative levels for a given dataset.

Another interesting baseline is a classifier that predicts shots randomly according to the relative frequency of shots to non-shots in the learning set, which would be the most trivial exploitation of the weak labels over a random predictor. While such a model would report a fairly accurate total count in identical benchmark situations, it would vary wildly on individual inputs. This is detected in the error rate: the weighed random model obtains $E = \mathbf{31.5\%}$ for the Minimi 5.56 and $E = \mathbf{37.5\%}$ for the M2. Note that since the class imbalance depends on the external conditions and usage, practical applications or different datasets are likely to have a different class distribution which would make such a model increasingly inaccurate. Such a model would also produce a large number of false positives during non-shot usage, which is highly undesirable.

In the experimental results of this section, we also mention (in smaller print) the raw counting result, which is the metric eventually used by the armourer and allows for some desirable error compensation. This also lets the reader know how many real shots an error rate is based on.
\subsubsection{Minimi 5.56} \label{Minimi556_experiments}
Table~\ref{tab:real_world_Minimi556} compares our method with DTC and the human-generated algorithm. Real-world testing (model \#1) was performed in similar conditions but with different weapons and sensors. Since then, we have found better values for hyperparameters $T_H$ and $T_L$, which significantly improved performance, and report the results of this model as "future" as it has not yet undergone real-world testing. The machine-learning model runs in 65ms/inference on our testing platform and the preprocessing runs in 22µs/sample. It uses 41kB of program memory and has peak usage of 11kB of RAM.

\begin{table}[!ht]
	\caption{Error rates and reported count of Minimi 5.56 algorithms on the validation set and during real-world testing, broken down by ammunition type.} \label{tab:real_world_Minimi556}
	\begin{tabular}
		{
			>{\centering}m{0.18\textwidth}
			|>{\centering}m{0.127\textwidth}
			>{\centering}m{0.127\textwidth}
			|>{\centering}m{0.127\textwidth}
			>{\centering}m{0.127\textwidth}
			>{\centering}m{0.127\textwidth}
			|>{\centering\arraybackslash}m{0.127\textwidth}
			>{\centering\arraybackslash}m{0.127\textwidth}
		}
		~ & \multicolumn{2}{c}{Human} & \multicolumn{3}{c}{EDGAR} & DTC \\
		~		               & valid. 	& real 			& valid. \scriptsize{(model~\#1)} 	& real \scriptsize{(model~\#1)}		& valid. \scriptsize{(future)} & valid. \\ \hline
		\multirow{2}{*}{Live}  & 2.07\%		& 0.41\%		& 0.17\%	& 2\% 		& 0\% 									  & 9.2\%	\\
		& {\tiny 1752/1785} 	& {\tiny 1205/1210}		& {\tiny 1788/1785} & {\tiny 404/400}   & {\tiny 1785/1785}           & {\tiny 1841/1785} \\
		\multirow{2}{*}{Blank} & 0.69\%    	& 0.74\%		& 0.28\%	& 0.75\%  	& 0\% 									  & 11.4\% \\
		& {\tiny 714/719}	& {\tiny 1201/1210}		& {\tiny 719/719} 	& {\tiny 403/400} 	& {\tiny 719/719}                   & {\tiny 795/719}\\  \hline
		\textbf{Total} 		   & \textbf{1.68\%} & \textbf{0.58\%} & \textbf{0.20\%} & \textbf{1.37\%}  & \textbf{0\%}		  & \textbf{9.82\%} \\ \hline
		Non-shot only			   %
		\scriptsize{(false positives)} 	   & $\nicefrac{30}{\text{\tiny>}55}$  & $\nicefrac{0}{\text{\tiny>}100}$    & $\nicefrac{0}{\text{\tiny>}55}$  & $\nicefrac{6}{\text{\tiny>}100}$  & $\nicefrac{0}{\text{\tiny>}55}$       & $\nicefrac{34}{\text{\tiny>}55}$
	\end{tabular}
\end{table}

We observe that the model significantly outperforms not only the unsupervised baseline but also the human-generated algorithm on the validation set and that the "future" model is even able to obtain a perfect score.
The expert model shows better performance in real-world testing on a previously unseen weapon and sensor. We believe this is due to the validation set over-representing extreme cases compared to the nominal-case real-world test. Our machine learning model shows a slightly degraded error rate, falling behind the expert model, but still remains well below the threshold for commercial viability ($E<5\%$). We assume that this difference is mainly due to an input distribution shift to which the machine learning model is sensitive. In particular, we discovered that our dataset for this weapon did not accurately reproduce the moment at which our accelerometer starts sampling in practical situations, leading to a slight difference in inputs for the first shot of a burst.

In addition, some false positives were detected. This is not a surprising result as this weapon is notoriously difficult to filter out false positives for, as attested by the result of the expert algorithm on the validation set. We believe that increasing their representation in the learning set would alleviate this issue.

\subsubsection{M2}
The M2 weapon platform provides a special firing mode allowing the user to manually control the weapon cycle, ensuring only single shots are possible. This functionality fundamentally alters the weapon cycle and, so far, human-generated algorithms have not been able to support it without suffering from a significant number of false-positives. %
These are currently completely ignored by existing shot counters. With this experiment, we have attempted to solve this problem by training and testing the neural network on the whole dataset, including 20\% of samples using this functionality. Results are compared in Table~\ref{tab:real_world_M2}. In addition, the M2 platform offered us the unique possibility of mounting two counters in parallel, thus providing directly comparable results.

The machine-learning model runs in 87ms/inference on our testing platform and the preprocessing runs in 22µs/sample. It uses 52kB of program memory and has peak usage of 14kB of RAM.

\begin{table}[!ht]
	\caption{Error rates and reported count of M2 algorithms on the validation set and during real-world testing, broken down by ammunition and mount types.} \label{tab:real_world_M2}
	\begin{tabular} %
		{
			>{\centering}m{0.1\textwidth}
			>{\centering}m{0.1\textwidth}
			|>{\centering}m{0.07\textwidth}
			>{\centering}m{0.07\textwidth}
			|>{\centering}m{0.07\textwidth}
			>{\centering}m{0.07\textwidth}
			|>{\centering}m{0.07\textwidth}
			||>{\centering}m{0.07\textwidth}
			>{\centering}m{0.07\textwidth}
			|>{\centering}m{0.07\textwidth}
			>{\centering}m{0.07\textwidth}
			|>{\centering\arraybackslash}m{0.07\textwidth}
		}
		~ && \multicolumn{5}{c}{Full-auto only} & \multicolumn{5}{c}{Full-auto + Manual rearm} \\ 
		\multirow{3}{*}{Fixation} & \multirow{3}{*}{Ammun.} & \multicolumn{2}{c}{Human} & \multicolumn{2}{c}{EDGAR} & DTC & \multicolumn{2}{c}{Human} & \multicolumn{2}{c}{EDGAR} & DTC \\
		~ && valid. & real & valid. & real & valid. & valid. & real & valid. & real & valid. \\ \hline
		\multirow{4}{*}{Tripod} 			 & \multirow{2}{*}{live}  & 0\% & 0\% & 0\%  & 12\% & 23\%  & 16\% & 20\% & 1\% & 10\% & 25\% \\
		&       & {\tiny 160/160} & {\tiny 40/40} & {\tiny 160/160} & {\tiny 35/40} & {\tiny 153/160} & {\tiny 160/190} & {\tiny 40/50} & {\tiny 188/190} & {\tiny 45/50} & {\tiny 193/190} \\
		& \multirow{2}{*}{blank} & 0\% & 0\% & 0\%  & 20\% & 38\%  & 20\% & 18\% & 2\% & 22\% & 62\% \\ %
		&       & {\tiny 160/160} & {\tiny 40/40} & {\tiny 160/160} & {\tiny 32/40} & {\tiny 159/160} & {\tiny 160/200} & {\tiny 40/49} & {\tiny 195/200} & {\tiny 38/49} & {\tiny 262/200} \\ \hline
		\multirow{4}{*}{Fixed} 		 	     & \multirow{2}{*}{live}  & 0\% & 0\% & 5\%  & 0\% & 58\% & 13\% & 20\% & 4\%  & 4\% & 56\% \\
		&       & {\tiny 209/209} & {\tiny 40/40} & {\tiny 203/209} & {\tiny 40/40} & {\tiny 275/209} & {\tiny 209/239} & {\tiny 40/50} & {\tiny 229/239} & {\tiny 48/50} & {\tiny 318/239} \\
		& \multirow{2}{*}{blank} & 0\% & 0\% & 2\%  & 17\% & 36\% & 24\% & 20\% & 1\%  & 26\% & 41\% \\ %
		&       & {\tiny 130/130} & {\tiny 40/40} & {\tiny 131/130} & {\tiny 33/40} & {\tiny 101/130} & {\tiny 130/170} & {\tiny 40/50} & {\tiny 169/170} & {\tiny 37/50} & {\tiny 159/170} \\ \hline
		\multirow{4}{*}{Elastic} 			 & \multirow{2}{*}{live}  & 0\% & 0\% & 1\%  & 0\% & 26\% & 18\% & 20\% & 1\%  & 10\% & 26\% \\
		&       & {\tiny 140/140} & {\tiny 40/40} & {\tiny 139/140} & {\tiny 40/40} & {\tiny 160/140} & {\tiny 140/170} & {\tiny 40/50} & {\tiny 169/170} & {\tiny 45/50} & {\tiny 185/170} \\
		& \multirow{2}{*}{blank} & 0\% & 6\%  & 6\%  & 7\% & 39\% & 19\% & 26\% & 5\%  & 10\% & 41\% \\ %
		&       & {\tiny 129/129} & {\tiny 75/80} & {\tiny 121/129} & {\tiny 37/40} & {\tiny 157/129} & {\tiny 129/159} & {\tiny 75/101} & {\tiny 151/159} & {\tiny 91/101} & {\tiny 182/159} \\ \hline
		\multirow{2}{*}{deFNder} 	         & \multirow{2}{*}{live}  & 0\% & 0\% & 0\% & 0\% & 30\% & 6\% & 22\% & 0\% & 8\% & 42\% \\ %
		&       & {\tiny 110/110} & {\tiny 39/39} & {\tiny 110/110} & {\tiny 40/40} & {\tiny 103/110} & {\tiny 122/130} & {\tiny 39/50} & {\tiny 130/130} & {\tiny 46/50} & {\tiny 144/130} \\ \hline
		\multicolumn{2}{c|}{\textbf{Average}} & \textbf{0\%} & \textbf{1\%} & \textbf{2\%} & \textbf{8\%} & \textbf{36\%} & \textbf{16\%} & \textbf{21\%} & \textbf{2\%} & \textbf{13\%} & \textbf{42\%} \\ \hline
		\multicolumn{2}{c|}{\makecell{Non-shot only\\\scriptsize{(false positives)}}} %
				& $\frac{0}{\text{\tiny>}110}$ & $\frac{0}{\text{\tiny>}80}$ & $\frac{0}{\text{\tiny>}180}$ & $\frac{0}{\text{\tiny>}80}$ & $\frac{314}{\text{\tiny>}180}$ & $\frac{0}{\text{\tiny>}180}$ & $\frac{0}{\text{\tiny>}80}$ & $\frac{0}{\text{\tiny>}180}$ & $\frac{0}{\text{\tiny>}80}$ & $\frac{314}{\text{\tiny>}180}$ \\
	\end{tabular}
\end{table}

Following the Minimi results, a greater emphasis was put on including non-shot data in the learning set. This seems to have proven successful as no false-positives have been detected during transport, setup, normal manipulations and over 80 supplementary manipulations and dry firings.

On the validation set, the human-generated algorithm obtains a perfect score on the partial dataset but rises to a 16\% error rate when including manually rearmed shots. The EDGAR model shows satisfactory performance ($E<5\%$) on the partial and full datasets by obtaining a stable 2\% error rate on both.

Due to the larger class imbalance of this dataset causing it to report many more false positives, the unsupervised baseline performs very poorly on this dataset and does not manage to beat the previously described weighed random model. This interpretation is supported by the non-shot data on which the DTC model reports (multiple) shots for every weapon manipulation, which would be unacceptable for practical use. We thus show a large improvement from this baseline. %

Real-world testing shows performance falling for both algorithms, especially the machine-learning model rising to 8\%. Results on the ammunition discrimination problem outside the scope of this paper suggest that this is at least partly due to a distribution shift in the input, which affects the machine learning model more significantly. Due to equipment availability constraints, only one weapon and three sensors were used to construct the dataset, while testing was done on brand-new, unseen weapons and sensors. Despite this, when including manually rearmed shots, the machine-learning model still shows overall better performance than the human-generated one.

We assume that supplementary testing on a more diverse batch of weapons and sensors could significantly improve the real-world performance of the model.

\autofigure[Effects of incremental improvements. Each box represents the error rate of 20 randomly initialized models on the validation set.]{width=\defaultFigWidth\textwidth}{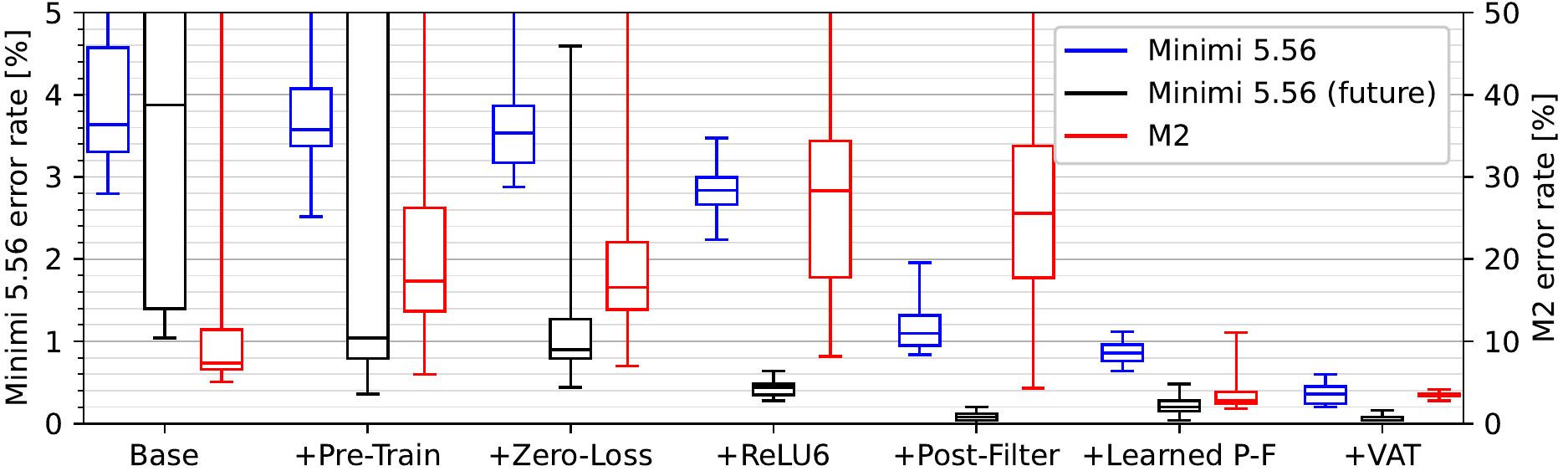}

\subsubsection{Breakdown of Performance Improvements} \label{performance_breakdown}
In Figure~\ref{fig:graph_paper_compare_improvements_merged_small}, we compare the performance of incremental improvements to the base method as discussed in Section~\ref{EDGAR}. We start with the base technique. As we move to the right we successively add improvements, conserving all previously enabled ones. Hyperparameters are identical to those used to obtain the experimental results in the previous sections. Some models fail to converge at all, especially when zero-loss is disabled; these are treated as 100\% error rate and explain the high maximums of some boxes.

We first enable pre-training, add our second term of Equation~\ref{eqn:lprop} in "Zero-Loss," and replace ReLU activations with ReLU6 (see  Section~\ref{model_structure}). We then enable simple post-filtering (see Section~\ref{minimum_cycle_time}), add our implementation of Algorithm~\ref{alg:prediction_mask} in "Learned P-F," and finally enable VAT (see Section~\ref{VAT}).%

\section{Conclusion and future work}
In this study, we have presented an approach that successfully learns an instance-level shot classifier from a weakly-labelled dataset, which we believe opens the way for new machine-learning applications. We showed that it significantly improves predictions from the unsupervised baseline by successfully exploiting the information from weak labels. We also showed that it outperforms human-generated algorithms in most offline testing, reach commercially acceptable levels of performance in real-world testing and are able to solve previously unanswered problems. We expect this to save several weeks of development time per product and enable increased customisation to specific platforms thus further increasing performance. Finally, we showed that our technique could yield models giving this level of performance in real-time on microcontrollers while using less than 14kB of RAM and 87ms per inference. In the future, we aim to bring real-world performance closer to the one obtained on offline benchmarks by improving the data collection technique and taking systematic measures against input distribution shifts. Outside the scope of this paper, our technique has also shown consistent better-than-human performance on the ammunition discrimination problem. It is already being deployed for that purpose in commercial applications. We hope to bring to light these results and the associate technique improvements in a further study.

\subsubsection{Acknowledgements} I would like to thank my colleague Louis Huggenberger for his continuous help in data acquisition. I would like to thank my thesis supervisors Hugues Libotte and Louis Wehenkel for their precious advice and suggestions.

\bibliographystyle{splncs04}
\bibliography{bibliography}{}
\end{document}